\documentclass[runningheads]{llncs}
\usepackage{graphicx}
\usepackage{comment} % for block comments
\usepackage{booktabs} % professional-quality tables
\usepackage{amssymb} % for math symbols
\usepackage{amsmath} % for math symbols
\usepackage{xcolor}

\usepackage[misc,geometry]{ifsym}

\begin{document}
\title{Neural Representation Learning for \\ Scribal Hands of Linear B}
%
%\titlerunning{Abbreviated paper title}
% If the paper title is too long for the running head, you can set
% an abbreviated paper title here
%

\author{Nikita Srivatsan\inst{1} \Letter \and
Jason Vega\inst{3} \and \\
Christina Skelton\inst{2} \and
Taylor Berg-Kirkpatrick\inst{3}}
\authorrunning{Srivatsan et al.}
% First names are abbreviated in the running head.
% If there are more than two authors, 'et al.' is used.
%
\institute{Language Technologies Institute, Carnegie Mellon University \\
\email{nsrivats@cmu.edu} \and
Center for Hellenic Studies, Harvard University \\
\email{cskelton@ucla.edu} \and
Computer Science and Engineering, University of California, San Diego \\
\email{\{jvega@,tberg@eng.\}ucsd.edu}}

\maketitle              % typeset the header of the contribution

\begin{abstract}

In this work, we present an investigation into the use of neural feature extraction in performing scribal hand analysis of the Linear B writing system.
While prior work has demonstrated the usefulness of strategies such as phylogenetic systematics in tracing Linear B's history, these approaches have relied on manually extracted features which can be very time consuming to define by hand.
Instead we propose learning features using a fully unsupervised neural network that does not require any human annotation.
Specifically our model assigns each glyph written by the same scribal hand a shared vector embedding to represent that author's stylistic patterns, and each glyph representing the same syllabic sign a shared vector embedding to represent the identifying shape of that character.
Thus the properties of each image in our dataset are represented as the combination of a scribe embedding and a sign embedding.
We train this model using both a reconstructive loss governed by a decoder that seeks to reproduce glyphs from their corresponding embeddings, and a discriminative loss which measures the model's ability to predict whether or not an embedding corresponds to a given image.
Among the key contributions of this work we (1) present a new dataset of Linear B glyphs, annotated by scribal hand and sign type, (2) propose a neural model for disentangling properties of scribal hands from glyph shape, and (3) quantitatively evaluate the learned embeddings on findplace prediction and similarity to manually extracted features, showing improvements over simpler baseline methods.

\end{abstract}

\section{Introduction}

Neural methods have seen much success in extracting high level information from visual representations of language.
Many tasks from OCR, to handwriting recognition, to font manifold learning have benefited greatly from relying on architectures imported from computer vision to learn complex features in an automated, end-to-end trainable manner.
However much of the emphasis of this field of work has been on well-supported modern languages, with significantly less attention paid towards low-resource, and in particular, historical scripts.
While self-supervised learning has seen success in these large scale settings, in this work we demonstrate its utility in a low data regime for an applied analysis task.
Specifically, we propose a novel neural framework for script analysis and present results on scribal hand representation learning for Linear B.

Linear B is a writing system that was used on Crete and the Greek mainland ca. 1400-1200 BCE.
It was used to write Mycenaean, the earliest dialect of Greek, and was written on leaf or page-shaped clay tablets for accounting purposes.
Linear B is a syllabary, with approximately $88$ syllabic signs as well as a number of ideograms, which are used to indicate the commodity represented by adjacent numerals.
The sites which have produced the most material to date are Knossos, Pylos, Thebes, and Mycenae (see Palmer~\cite{palmer2008begin} for more detail) but for the purposes of this paper, we focus specifically on Knossos and Pylos from which we collect a dataset of images of specific instances of glyphs written by $74$ different scribal hands.
Many signs exhibit slight variations depending on which scribe wrote them.
Therefore, being able to uncover patterns in how different scribes may write the same character can inform us about the scribes' potential connections with one another, and to an extent even the ways in which the writing system evolved over time.
This is a process that has previously been performed by hand~\cite{skelton2008methods}, but to which we believe neural methods can provide new insights.

Rather than building representations of each scribe's writing style based on the presence of sign variations as determined by a human annotator, we propose a novel neural model that disentangles features of glyphs relating to sign shape and scribal idiosyncrasies directly from the raw images.
Our model accomplishes this by modeling each image using two separate real-valued vector embeddings --- we share one of these embeddings across glyphs written by the same scribe, and the other across glyphs depicting the same sign, thereby encouraging them to capture relevant patterns.
These embeddings are learned via three networks.
The first is a decoder that attempts to reconstruct images of glyphs from their corresponding embeddings using a series of transpose convolutional layers.
We penalize the reconstructed output based on its mean squared error (MSE) vs. the original.
In addition, we also use two discriminator networks, which attempt to predict whether a given image-embedding pair actually correspond to one another or not, both for scribe and sign embeddings respectively.
These losses are backpropagated into those original vectors to encourage them to capture distinguishing properties that we care about.

This model has many potential downstream applications, as paleography has long been an integral part of the study of the Linear B texts~\cite{palaima2011scribes}.  For a handful of examples from prior work, Bennett's study of the Pylos tablets~\cite{bennett1947minoan} revealed the existence of a number of different scribal hands, all identified on the basis of handwriting, since scribes did not sign their work.  This research was carried out even before the decipherment of Linear B.  Later, Driessen~\cite{driessen2000scribes} used the archaic sign forms on Linear B material from the Room of the Chariot Tablets at Knossos, in conjunction with archaeological evidence, to argue that these tablets pre-dated the remainder of the Linear B tablets from Knossos.  Skelton~\cite{skelton2008methods} and later Skelton and Firth~\cite{skelton2016cstudy} carried out a phylogenetic analysis of Linear B based on discrete paleographical characteristics of the sign forms, and were able to use this analysis to help date the remainder of the Linear B tablets from Knossos, whose relative and absolute dates had already been in question.
Our hope is that leveraging the large capacity of neural architectures can eventually lead to further strides in these areas.

Additionally, qualitative assessments of Linear B paleography have long been used to date tablets whose history or findplaces are uncertain.
This has been vital in the case of the Knossos tablets, which were excavated before archaeological dating methods were well-developed~\cite{firth2016astudy}, and the tablets from the Pylos megaron, which seem to date to an earlier time period than the rest of the palace~\cite{skelton2011look}.
While qualitative techniques have been fruitful, a key motivating question in our work is to what extent we can use quantitative methods to improve on these results.

Going back further, Linear B was adapted from the earlier writing system Linear A, which remains undeciphered.
There is a hundred-year gap between the last attestation of Linear A and the first attestation of Linear B.
With a more thorough understanding of the evolution of these two writing systems, can we reconstruct the sign forms of the earliest Linear B, and elucidate the circumstances behind its invention?
With this in mind, our approach explicitly incorporates a decoder network capable of reconstructing glyphs from learned vector representations, that we hope can eventually be useful for such investigation.

In total, three writing systems related to Linear B remain undeciphered (Cretan Hieroglypic, Linear A, and Cypro-Minoan), and a fourth (the Cypriot Syllabary) was used to write an unknown language.
One motivating question is whether a better understanding of the evolution of sign forms would help contribute additional context that could make a decipherment possible.
In particular, scholars are not yet in agreement over the sign list for Cypro-Minoan, and how many distinct languages and writing systems it may represent~\cite{ferrara2012cypro}.
Understanding variation in sign forms is foundational for this work~\cite{palaima2011scribes}, and it makes sense to test our methods on a related known writing system (i.e. Linear B) before attempting a writing system where so much has yet to be established.

Overall these are just some of the applications that we hope the steps presented here can be directed towards.

\section{Prior Work}

\begin{figure}
    \centering
    \includegraphics[width=0.41\textwidth]{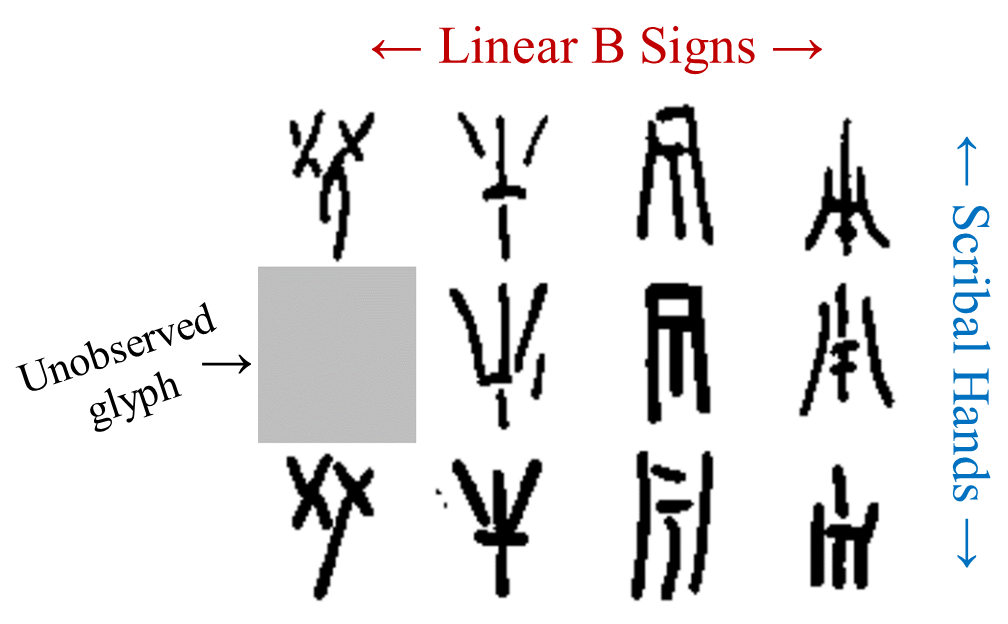}
    \caption{Binarized examples from our collected dataset. Note the stylistic differences particularly in the left and right most signs, and that we do not have an observation for the left most sign in the middle scribe's hand.}
    \label{fig:data}
\end{figure}

One of our primary points of comparison is the work of Skelton~\cite{skelton2008methods}, who manually defined various types of systematic stylistic variations in the way signs were written by different scribal hands, and used these features to analyze similarity between scribes via phylogenetic systematics.
In a sense this work can serve as our human baseline, giving insight into what we might expect reasonable output from our system to look like, and helping us gauge performance in an unsupervised domain without a clear notion of ground truth.

While neural methods have not previously been used in analyzing this particular writing system, there is established literature on applying them to typography more broadly.
For example, Srivatsan et al.~\cite{srivatsan2019deep} used a similar shared embedding structure to learn a manifold over font styles and perform font reconstruction.
There has also been extensive work on writer identification from handwritten samples~\cite{bulacu2007text,siddiqi2010text,zhang2016end,djeddi2018icfhr}, as well as writer retrieval~\cite{christlein2019icdar}.
However these approaches rely on supervised discriminative training, in contrast to ours which is not explicitly trained on predicting authorship, but rather learning features.

\begin{figure}
    \centering
    \includegraphics[width=\textwidth]{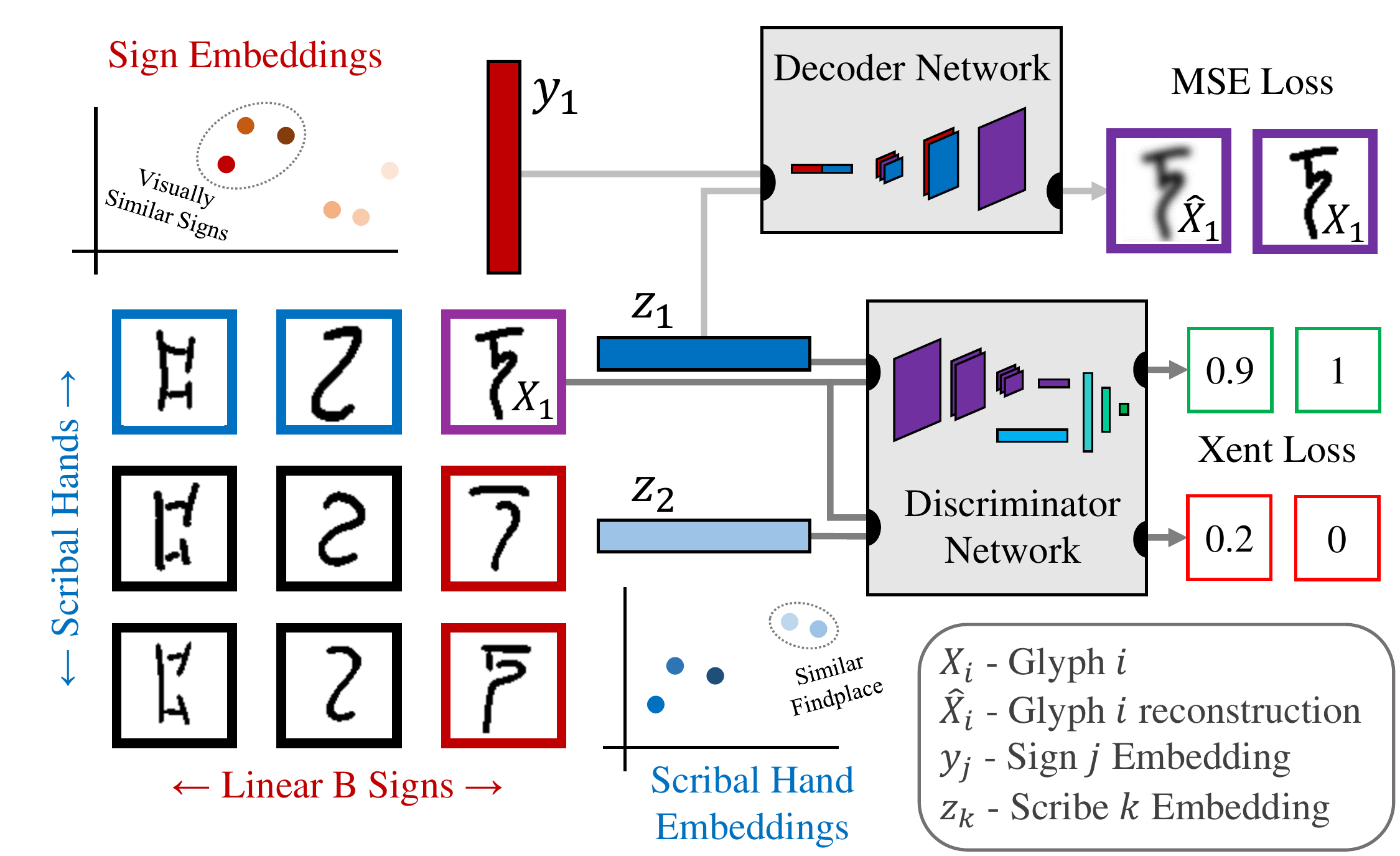}
    \caption{Diagram of the model --- each row is assigned a scribe embedding, and each column is assigned a sign embedding, which feed into a decoder and discriminator to collectively compute the loss function.}
    \label{fig:model}
\end{figure}

\section{Methods}

In this section we describe the layout of our model and its training procedure.
At a high level, our goal is to train it on our dataset of images of glyphs, and in doing so learn vector embeddings for each scribe that encode properties of their stylistic idiosyncrasies --- for example which particular variations of each sign they preferred to use --- while ignoring more surface level information such as the overall visual silhouettes of the signs we happen to observe written by them.
To do this, we break the roles of modeling character shape and scribal stylistic tendencies into separate model parameters, as we now describe.

Our model, depicted in Figure~\ref{fig:model}, assigns each sign and each scribal hand a vector embedding to capture the structural and stylistic properties of glyphs in our data.
These embeddings are shared by glyphs with the same sign or scribe, effectively factorizing our dataset into a set of vectors, one for each row and column.
The loss is broken into two parts, a reconstruction loss and a discriminative loss.
The reconstruction loss is computed by feeding a pair of sign and scribe embeddings to a decoder, which then outputs a reconstruction of the original glyph that those embeddings correspond to.
A simple MSE is computed between this and the gold image.
The discriminative cross-entropy loss is computed by the discriminator, a binary classifier which given an image of a glyph and a scribe embedding must predict whether or not the two match (i.e. did that scribe write that glyph).
We train this on equally balanced positive and negative pairs.
A similar discriminator is applied to the sign embeddings.
In our experiments, we ablate these three losses and examine the effects on performance.

More formally, suppose we have a corpus of images of glyphs $X$, where $X_i \in [0,1]^{64 \times 64}$ depicts one of $J$ characters and is authored by one of $K$ scribes.
We define a set of vectors $Y$, where each $Y_j \in \mathbb{R}^{d}$ represents features of sign $j$.
Similarly we define a set of vectors $Z$, where each $Z_k \in \mathbb{R}^{d}$ represents features of scribe $k$.
Thus, a given $X_i$ depicting sign $j$ and authored by scribe $k$ is represented at a high level in terms of its corresponding $Y_j$ and $Z_k$, each of which is shared by other images corresponding to the same character or scribal hand respectively.

For a given $X_i$, the decoder network takes in the corresponding pair of vectors $Y_j$ and $Z_k$, and then passes them through a series of transpose convolutional layers to produce a reconstructed image $\hat{X}_i$.
This reconstruction incurs a loss on the model based on its MSE compared to the original image.

Next, the scribal hand discriminator with parameters $\phi$ passes $X_i$ through several convolutional layers to obtain a feature embedding which is then concatenated with the corresponding $Z_k$ and passes through a series of fully connected layers, ultimately outputting a predicted $p(Z_k | X_i ; \phi)$ which scores the likelihood of the scribal hand represented by $Z_k$ being the true author of $X_i$.
Similarly, a sign discriminator with parameters $\psi$ predicts $p(Y_j | X_i ; \psi)$, i.e. the likelihood of $X_i$ depicting the sign represented by $Y_j$.
These discriminators are shown both one true match and one false match for an $X_i$ at every training step, giving us a basic cross-entropy loss.

These components all together yield the following loss function for our model:
\begin{equation*}
\begin{split}
    L(X_i,Y_j,Z_k) =& \lVert X_i - \hat{X}_i \rVert_2 \\
    +& \lambda_1 * ( \log p(Y_j | X_i ; \phi) + \log (1 - p(Y_{j'} | X_i ; \phi)) ) \\
    +& \lambda_2 * ( \log p(Z_k | X_i ; \psi) + \log (1 - p(Z_{k'} | X_i ; \psi)) ) \\
\end{split}
\end{equation*}
where $j'$ and $k'$ are randomly selected such that $j \neq j'$ and $k \neq k'$.

The format of this model lends itself out of the box to various downstream tasks.
For example, when a new clay tablet is discovered, it may be useful to determine if the author was a known scribal hand or someone completely new.
The discriminator of our model can be used for exactly such as a purpose, as it can be directly queried for the likelihood of a new glyph matching each existing scribe.
Even if this method is not exact enough to provide a definitive answer, we can at least use the predictions to get a rough sense of who the most similar known scribes are, despite never having seen that tablet during training.

Also, while our model treats the scribe and sign embeddings as parameters to be directly learned, one could easily imagine an extension in which embeddings are inferred from the images using an encoder network.
Taking this further, we could even treat those embeddings as latent variables by adding a probabilistic prior, such as in Variational Autoencoders~\cite{vae}.
This would let us compute embeddings without having trained on that scribe's writing, taking the idea from the previous paragraph one step further, and letting us perform reconstructions of what we expect a full set of signs by that new scribe to look like.

\subsection{Architecture}

We now describe the architecture of our model in more detail.
Let $F_i$ represent a fully connected layer of size $i$, $T$ represent a $2$x$2$ transpose convolutional layer, $C$ be a $3$x$3$ convolution, $I$ be an instance normalization layer~\cite{instancenorm}, $R$ be a ReLU, $M$ be a $2$x$2$ blur pool~\cite{zhang2019making}.
The decoder network is then $F_{1024*4*4} \rightarrow 4 \times ( T \rightarrow 2 \times ( C \rightarrow I \rightarrow R ) )$, where each convolutional layer $C$ reduces the number of filters by half.
The discriminator's encoder network is then $C \rightarrow 4 \times ( 3 \times ( C \rightarrow R ) \rightarrow M ) \rightarrow F_{16}$, where each convolutional layer $C$ increases the number of filters by double, except for the first which increases it to $64$.
Having encoded the input image, the discriminator concatenates the output of its encoder with the input scribe/sign embedding, and then passes the result through $7$ fully connected layers to obtain the final output probability.

\section{Data}

\begin{figure}
    \centering
    \includegraphics[width=0.9\textwidth]{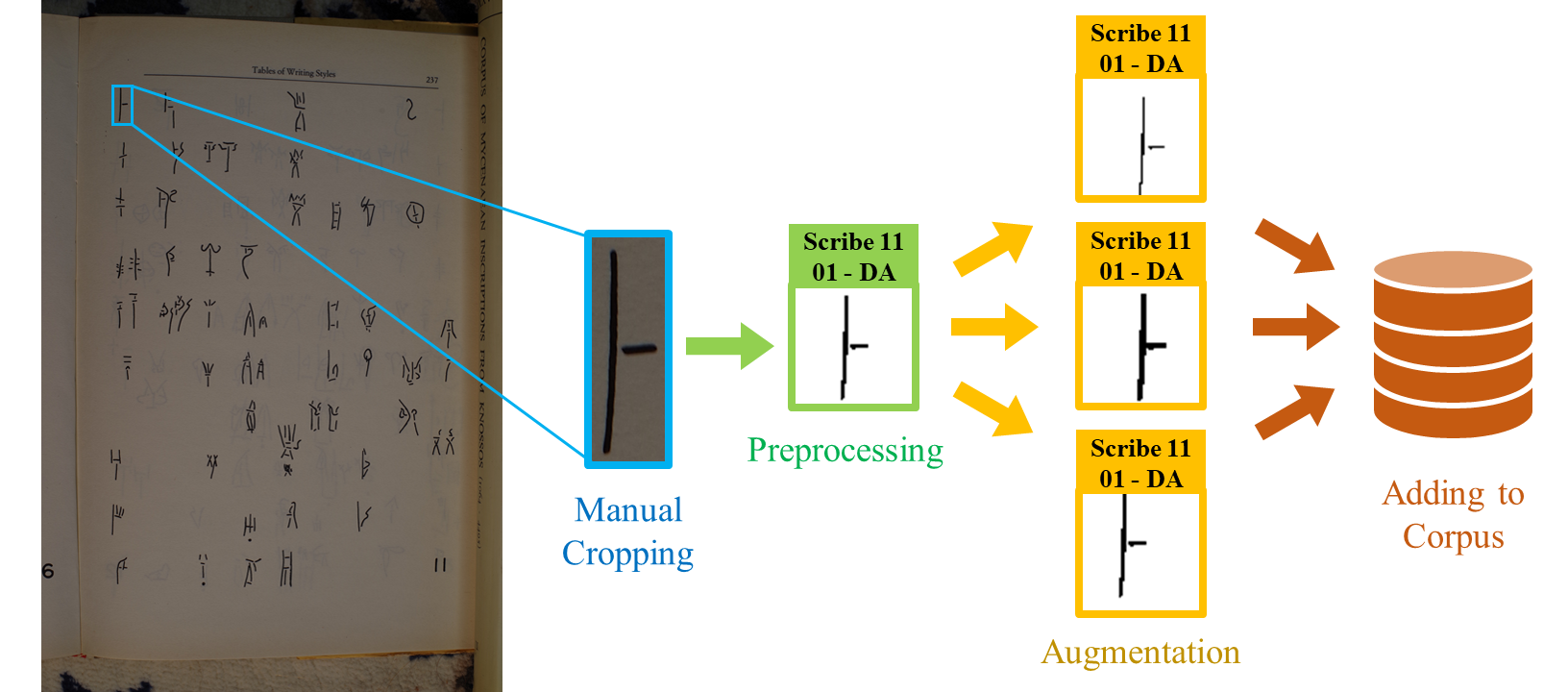}
    \caption{Visualization of the data pipeline. Glyphs are (1) manually cropped from photos of pages and labeled by sign and scribe, (2) preprocessed to $64$x$64$ binary images, and (3) augmented to produce multiple variants, before being added to the overall corpus.}
    \label{fig:cropping}
\end{figure}

\subsection{Collection}

There has been previous work on transcribing the tablets of Knossos by Olivier~\cite{olivier1965scribes} and Pylos by Palaima~\cite{palaima1988scribes}.
These authors present their transcriptions in form of a rough table for each hand containing every example of each sign written by them.
While these are an important contribution and certainly parseable by a human reader, they are not of a format readily acceptable by computer vision methods.
Since our model considers the atomic datapoint to be an individual glyph, we must first crop out every glyph from each table associated with a scribal hand.
This is a nontrivial task as the existing tables are not strictly grid aligned, and many glyphs overlap with one another, which requires us to manually paint out the encroaching neighbors for many images.
Ultimately we were able to convert the existing transcriptions into a digital corpus of images of individual glyphs, labeled both by scribal hand as well as sign type.
This process was carried out by two graduate students under the guidance of a domain expert.
Before being fed into our model, images were also binarized via Otsu's thresholding~\cite{otsu1979threshold}.

The digitization of this dataset into a standardized, well-formatted corpus is of much potential scientific value.
While this process was laborious, it opens the doors not just to our own work, but also to any other future research hoping to apply machine learning methods to this data.
Of course had this been a high-resource language such as English, we would have been able to make use of the multitude of available OCR toolkits to automate this preprocessing step.
As such, one possible direction for future work is to train a system for glyph detection and segmentation, for which the corpus we have already created can serve as a source of supervision.
This would be useful in more efficiently cropping similarly transcribed data, whether in Linear B or other related languages.

\subsection{Augmentation}

% image of an augmented glyph
\begin{figure}
    \centering
    \includegraphics[width=0.6\textwidth]{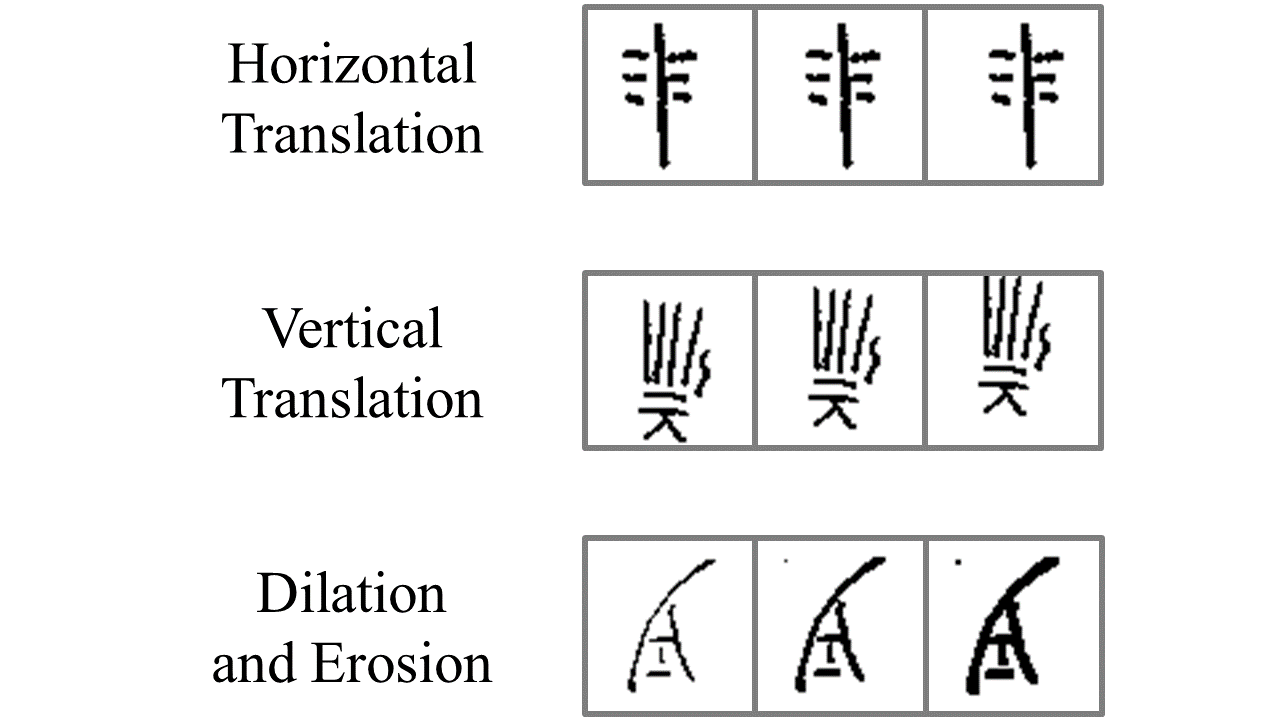}
    \caption{Examples of the three types of data augmentation we perform. Each original image is augmented in all possible ways, resulting in an effective $27$x increase in the size of our corpus.}
    \label{fig:augmentation}
\end{figure}

Due to the limited size of this corpus, there is a serious risk of models overfitting to spurious artifacts in the images rather than properties of the higher level structure of the glyphs.
To mitigate this, we perform a data augmentation procedure to artificially inflate the number of glyphs by generating multiple variants of each original image that differ in ways we wish for our model to ignore.
These are designed to mimic forms of variation that naturally occur in our data --- for example, glyphs may not always be perfectly centered within the image due to human error during cropping, and strokes may appear thicker or thinner in some images based on the physical size of the sign being drawn relative to the stylus' width, or even as a result of inconsistencies in binarization during preprocessing.
Following these observations, the specific types of augmentation we use include translation by $5$ pixels in either direction horizontally and/or vertically, as well as eroding or dilating with a $2$x$2$ pixel wide kernel (examples shown in Figure~\ref{fig:augmentation}).
All together we produce $27$ variations of each image.
Doing this theoretically encourages our model to learn features that are invariant to these types of visual differences, and therefore more likely to capture the coarser stylistic properties we wish to study.

\subsection{Statistics}

Our dataset represents $88$ Linear B sign types, written by $74$ scribal hands (of these $48$ are from Knossos and $26$ are from Pylos).
For each scribal hand, we observe an average of $56$ glyph images, some which may represent the same sign.
Conversely each sign appears an average of $47$ times across all scribal hands.
This gives us a total of $4,134$ images in our corpus, which after performing data augmentation increases to $111,618$.

\section{Experiments}

Having trained our model, we need to be able to quantitatively evaluate the extent to which it has learned features reflective of the sorts of stylistic patterns we care about, something which is nontrivial given the lack of any supervised ground truth success criteria.
There are two metrics we use to accomplish this.
First, given that we expect scribes of the same findplace to share stylistic commonalities, we train a findplace classifier based on our scribal hand embeddings to probe whether our model's embeddings contain information that is informative for this downstream task.
Secondly, since we have human generated scribal hand representations in the form of Skelton's manual features~\cite{skelton2008methods}, we can measure the correlation between them and our model's learned manifold to see how well our embeddings align with human judgements of relevant features.
We now describe these evaluation setups in more detail.

\subsection{Findplace Classification}

The first way we evaluate our model is by probing the learned embeddings to see how predictive they are of findplace, a metadata attribute not observed during training.
Since our goal is to capture salient information about broad similarities between scribal hands, one coarse measure of our success at this is to inspect whether our model learns similar embeddings for scribes of the same findplace, as they would likely share similar stylistic properties.
One reason for this underlying belief is that such scribes would have been contemporaries of each other, as can be infered from the fact that existing Linear B tablets were written ``days, weeks, or at most months'' before the destruction of the palaces~\cite{palaima2011scribes}.
For any scribes working together at the same place and time, we observe similar writing styles, likely as a result of the way in which they would have been trained.
The number of scribes attested at any one palatial site is quite low, which would seem to rule out a sort of formal scribal school as attested in contemporary Near Eastern sources, and instead suggests that scribal training took place as more of a sort of apprenticeship, or as a profession passed from parents to children, as is attested for various sorts of craftsmen in the Linear B archives~\cite{palaima2011scribes}.
Thus, we expect that scribes found at the same findplace would exhibit similar writing styles, and therefore measuring how predictive our model's embeddings are of findplace serves as a reasonable proxy for how well they encode stylistic patterns.

To this end, we train a supervised logistic regression classifier (one per model) to predict the corresponding findplaces from the learned embeddings, and evaluate its performance on a held out set.
In principal, the more faithfully a model's embeddings capture distinguishing properties of the scribal hands, the more accurately the classifier can predict a scribe's findplace based solely on those neural features.
We perform $k=5$ fold cross validation, and report average $F_1$ score over the $5$ train/test splits.
Our final numbers reflect the best score over $15$ random initializations of the classifier parameters for each split.

Findplaces for the Knossos tablets were described in Melena and Firth~\cite{melena2019knossos}, and for Pylos in Palaima~\cite{palaima1988scribes}.
However, there are some findplaces from which we only have one or two scribal hands in our corpus.
As supervised classification is not practically possible in such cases, we exclude any findplaces corresponding to $2$ or fewer scribal hands from our probing experiments.
This leaves us with $63$ scribal hands, labeled by $8$ different findplaces to evaluate on.

\subsection{QVEC}

While findplace prediction is a useful downstream metric, we also wish to measure how well the neural features our model learns correlate with those extracted by expert humans.
In order to do this, we use \texttt{QVEC}~\cite{qvec} --- a metric originally developed for comparing neural word vectors with manual linguistic features --- to score the similarity between our learned scribal hand embeddings and the manual features of Skelton~\cite{skelton2008methods}.
At a high level, \texttt{QVEC} works by finding the optimal one-to-many alignment between the dimensions of a neural embedding space and the dimensions of a manual feature representation, and then for each of the aligned neural and manual dimensions, summing their Pearson correlation coefficient with respect to the vocabulary (or in our case the scribal hands).
Therefore, this metric provides an intuitive way for us to quantitatively evaluate the extent to which our learned neural features correlate with those extracted by human experts.
As a side note, while the original \texttt{QVEC} algorithm assumes a one-to-many alignment between neural dimensions and manual features, we implement it as a many-to-one, owing to the fact that in our setting we have significantly more manual features than dimensions in our learned embeddings, implying that each neural dimension is potentially responsible for capturing information about multiple manual features rather than vice versa as is the case in the original setting of Tsvetkov et al.~\cite{qvec}.

\subsection{Baselines}

Our primary baseline is a naive autoencoder, that assumes a single separate embedding for every image in the dataset, therefore not sharing embeddings across signs or scribal hands.
Its encoder layout is the same as that of our discriminator, and its decoder is identical to that of our main model so as to reduce differences due to architecture capacity.
In order to obtain an embedding for a scribal hand using this autoencoder, we simply average the embeddings of all glyphs produced by that scribe.
We report several ablations of our model, indicating the presence of the reconstruction loss, scribal hand discriminator loss, and sign discriminator loss by +Recon, +Scribe, and +Sign respectively.
Finally, we provide the $F_1$ score for a naive model that always predicts the most common class to indicate the lowest possible performance.

\subsection{Training Details}

We use sign and scribe embeddings of size $16$ each for all experiments.
Our model is trained with the Adam~\cite{adam} optimization algorithm at a learning rate of $10^{-4}$ and a batch size of $25$.
It is implemented in Pytorch~\cite{pytorch} version \texttt{1.8.1} and trains on a single NVIDIA 2080ti in roughly one day.
The classifier used for evaluation is trained via SGD at a learning rate of $10^{-3}$ and a batch size of $15$.

\section{Results}

% probing
\begin{table}
\centering
\caption{(Left) Best cross validated $F_1$ achieved by logistic regression findplace classifier on embeddings produced by ablations of our model, compared to an autoencoder, and naive most common label baseline. (Right) \texttt{QVEC}~\cite{qvec} score between models' embeddings and manual features of Skelton~\cite{skelton2008methods}.}
\begin{tabular}{lcc}
\toprule
\textbf{Model} & \textbf{Findplace} & \texttt{QVEC} \textbf{Similarity} \\
& \textbf{Classifier} $F_1$ & \textbf{to Manual}  \\
\midrule
Most Common & 0.154 & - \\
\midrule
Autoencoder & 0.234 & 54.1 \\
\midrule
-Recon, +Scribe, +Sign & 0.198 & 53.2 \\
+Recon, -Scribe, -Sign & 0.234 & 51.1 \\
+Recon, +Scribe, -Sign & 0.206 & 52.9 \\
+Recon, +Scribe, +Sign & \textbf{0.292} & \textbf{57.0} \\
\bottomrule
\end{tabular}
\label{tab:results}
\end{table}

\begin{comment}
% l2
\begin{table}
\centering
\caption{Average inter-cluster distance for embeddings produced by ablations of our model, as well as a random Gaussian baseline, and manual features from Skelton~\cite{skelton2008methods}.}
\begin{tabular}{rc}
\toprule
Model & Avg. Inter-Cluster Dist \\
\midrule
Random & 16.33 \\
\midrule
+Recon, -Scribe, -Sign & \textbf{15.87} \\
-Recon, +Scribe, -Sign & 16.12 \\
+Recon, +Scribe, -Sign & 15.95 \\
+Recon, +Scribe, +Sign & 16.20 \\
\midrule
Manual Features & \textbf{11.64} \\
\bottomrule
\end{tabular}
\label{tab:ablations}
\end{table}
\end{comment}

\begin{comment}
\begin{figure}
    \centering
    \includegraphics[width=\textwidth]{images/tsne.png}
    \caption{t-SNE projection of learned scribal hand embeddings, colored by findplace}
    \label{fig:tsne}
\end{figure}
\end{comment}

% f1 discussion
Table~\ref{tab:results} shows results of a findplace classifier trained on our system's output vs. that of an autoencoder, with the score for a most common baseline as a lower bound.
We see that our full model with all three losses achieves the highest score, with reduced performance as they are successively removed.
The worst score comes from the model trained on purely discriminative losses with no decoder, suggesting that the reconstruction loss provides important bias to prevent overfitting given the limited dataset size.
This also makes sense given the strong performance by the vanilla autoencoder.

For reference, Skelton~\cite{skelton2008methods} provides manual feature representations for $20$ of the scribal hands in our corpus ($17$ of which are from the $6$ findplaces shared by at least one other scribe in that set).
If we perform a similar probing experiment with the hand crafted features for those $17$ scribes, we get an $F_1$ score of $0.37$.
While this result is useful as an indicator of where the upper bound on this task may be, it is very important to note that this experiment is only possible on a small subset of the scribes we evaluate our model on and is therefore not directly comparable to the results in Table~\ref{tab:results}.
A more direct comparison to the neural features of our model would require extracting manual features for the remaining $54$ scribal hands, which is beyond the scope of this work.

% qvec discussion
Table~\ref{tab:results} also shows the \texttt{QVEC} similarity between each of the models' learned embeddings and the manual features of Skelton~\cite{skelton2008methods}.
Once again, our system ranks highest, with the autoencoder also performing more strongly than our ablations.
Interestingly, we see here that our discriminative losses appear more important than the reconstructive loss, suggesting that by encouraging the model to focus on features that are highly distinguishing between scribes or signs we may end up recovering more of the same information as humans would.

\begin{comment}
% qvec
\begin{table}
\centering
\caption{\texttt{QVEC}~\cite{qvec} score between various models and manual features of Skelton~\cite{skelton2008methods}.}
\begin{tabular}{lr}
\toprule
\textbf{Model} & \texttt{QVEC} \\
\midrule
Auto Encoder & 54.1 \\
\midrule
-Recon, +Scribe, +Sign & 53.2 \\
+Recon, -Scribe, -Sign & 51.1 \\
+Recon, +Scribe, -Sign & 52.9 \\
+Recon, +Scribe, +Sign & \textbf{57.0} \\
\bottomrule
\end{tabular}
\label{tab:qvec}
\end{table}
\end{comment}

\begin{figure}
    \centering
    \includegraphics[width=0.3\textwidth]{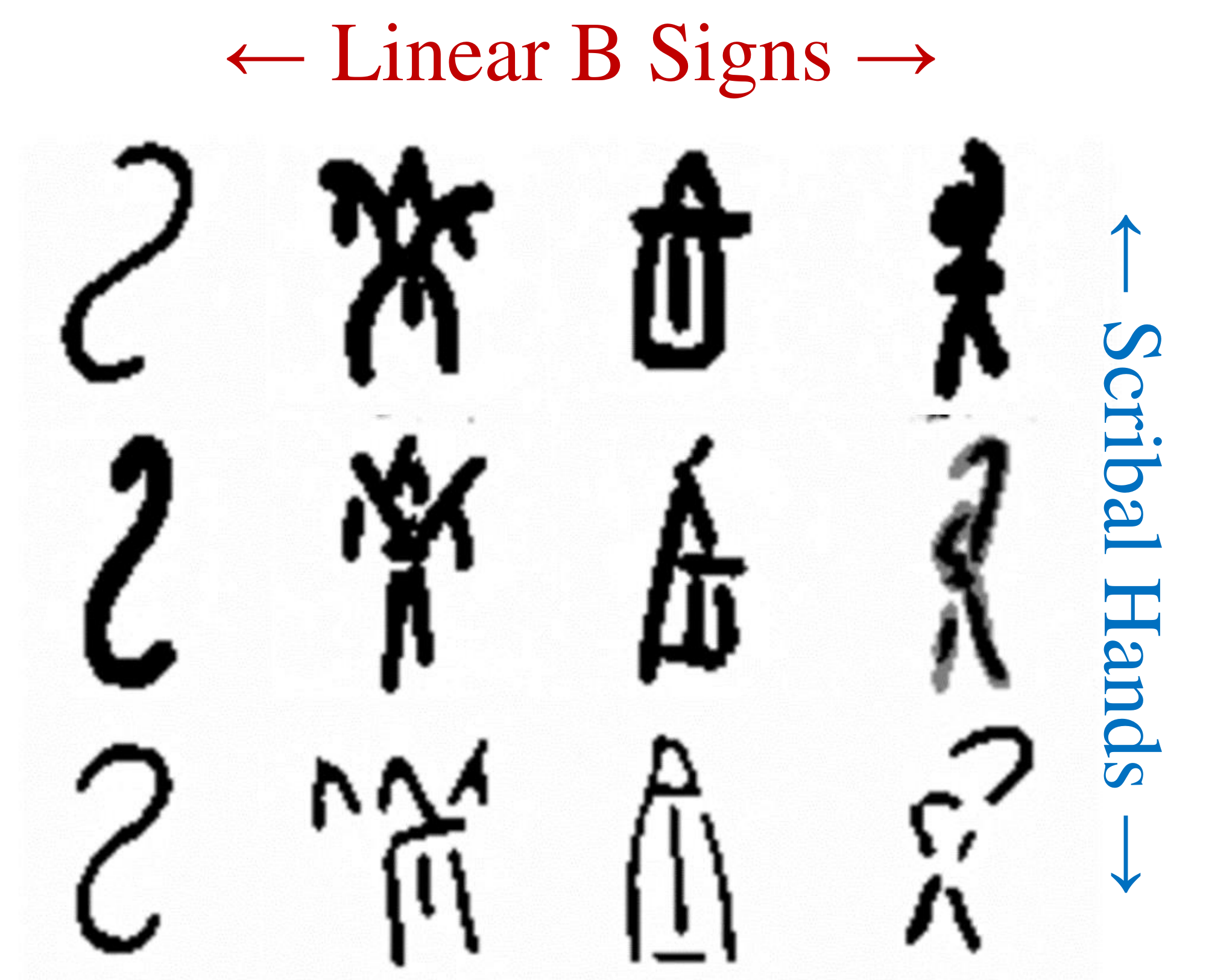}
    \caption{Example reconstructions from our model for various scribal hands. The model manages to capture subtle stylistic differences, but we see a failure mode in the middle row right column where the output appears as a superposition of two images.}
    \label{fig:recon}
\end{figure}

Figure~\ref{fig:recon} shows example reconstructions from our decoder.
These reconstructions are based solely on the scribe and sign representations for the corresponding row and column.
We see that the model is able to realistically reconstruct the glyphs, showing that it is very much capable of learning shared representations for sign shape and scribal hand styles, which can then be recombined to render back the original images.
This would imply that our embedding space is able to adequately fit the corpus, and perhaps requires some additional inductive bias in order to encourage those features to be more high level.

There are also additional downstream tasks we have not yet evaluated on that may be worth attempting with our system.
The most obvious is that of phylogeny reconstruction.
Now that we have trained scribal hand embeddings, it remains to be seen how similar a phylogenetic reconstruction based on these features would resemble that found by Skelton~\cite{skelton2008methods}.
This may give us a better idea as to whether the embeddings we have computed are informative as to the broader hierarchical patterns in the evolution of Linear B as a writing system.

%While these results are not yet necessarily conclusive, we believe that they do at least indicate that our model is on some level able to capture properties of scribal hands which are not uncorrelated with at least one held out indicator of coarse stylistic similarity.
%There's certainly more investigation to be done, but these early experiments do seem to confirm that such future work is warranted.

\section{Conclusion}
In this paper we introduced a novel approach to learning representations of scribal hands in Linear B using a new neural model which we described.
To that end, we digitized a corpus of glyphs, creating a more standardized dataset more readily amenable to machine learning methods going forwards.
We benchmarked the performance of this approach compared to simple baselines and ablations by quantitatively measuring how predictive the learned embeddings were of held out metadata regarding their findplaces.
In addition, we directly evaluated the similarity of our scribal hand representations to those from existing manual features.
We believe this work leaves many open avenues for future research, both with regard to the model design itself, as well as to further tasks it can be applied towards.
For example, one of the key applications of manual features in the past has been phylogenetic analysis, something which our neural features could potentially augment for further improvements.
This type of downstream work can further be used in service of dating Linear B tablets, understanding the patterns in the writing system's evolution, building OCR systems, or even analyzing other related but as of yet undeciphered scripts.

%
% ---- Bibliography ----
%
% BibTeX users should specify bibliography style 'splncs04'.
% References will then be sorted and formatted in the correct style.
%
\bibliographystyle{splncs04}
\bibliography{custom}

\end{document}